\documentclass[11pt]{article}
\usepackage{authblk}
\usepackage{eacl2017}
\usepackage{times}
\usepackage{url}
\usepackage{latexsym}
\usepackage{graphicx}
\usepackage{amsmath,amsfonts,amssymb}
\usepackage{bm,bbm}
\usepackage{array}
\usepackage{booktabs}
\usepackage{todonotes}

\eaclfinalcopy % Uncomment this line for the final submission
\def\eaclpaperid{337} %  Enter the acl Paper ID here

\title{Negative Sampling Improves Hypernymy Extraction\\Based on Projection Learning}

\author[$\dag$]{\bf Dmitry Ustalov}
\author[$\S$]{\bf Nikolay Arefyev}
\author[$\ddag$]{\bf Chris Biemann}
\author[$\ddag$]{\bf Alexander Panchenko}
\affil[$\dag$]{Ural Federal University, Institute of Natural Sciences and Mathematics, Russia}
\affil[$\S$]{Moscow State University, Faculty of Computational Mathematics and Cybernetics, Russia}
\affil[$\ddag$]{University of Hamburg, Deptartment of Informatics, Language Technology Group, Germany}
\affil[ ]{ \tt  dmitry.ustalov@urfu.ru, narefjev@cs.msu.ru}
\affil[  ]{ \tt  \{biemann,panchenko\}@informatik.uni-hamburg.de}

\date{}

\renewcommand{\vec}[1]{\ensuremath\bm{#1}}
\newcommand{\hitk}[1]{\ensuremath\text{hit@}{#1}}
\newcommand{\auc}{\ensuremath\operatorname{\text{AUC}}}
\newcommand{\nn}[1]{\ensuremath\operatorname{\text{NN}}_{#1}}

\begin{document}

\abovedisplayskip=4pt
\abovedisplayshortskip=1pt
\belowdisplayskip=1pt
\belowdisplayshortskip=1pt

\maketitle

\begin{abstract}
We present a new approach to extraction of hypernyms based on projection learning and word embeddings. In contrast to classification-based approaches, projection-based methods require no candidate hyponym-hypernym pairs. While it is natural to use both positive and negative training examples in supervised relation extraction, the impact of negative examples on hypernym prediction was not studied so far. In this paper, we show that explicit negative examples used for regularization of the model significantly improve performance compared to the state-of-the-art approach of~\newcite{Fu:14} on three datasets from different languages.
\end{abstract}

\section{Introduction}

Hypernyms are useful in many natural language processing tasks ranging from construction of taxonomies~\cite{Snow:06,panchenko-EtAl:2016:SemEval} to query expansion~\cite{Gong:05} and question answering~\cite{Zhou:13}. Automatic extraction of hypernyms from text has been an active area of research since manually constructed high-quality resources featuring hypernyms, such as WordNet~\cite{Miller:95}, are not available for many domain-language pairs.

The drawback of pattern-based approaches to hypernymy extraction~\cite{Hearst:92} is their sparsity. Approaches that rely on the classification of pairs of word embeddings~\cite{Levy:15} aim to tackle this shortcoming, but they require candidate hyponym-hypernym pairs. We explore a hypernymy extraction approach that requires no candidate pairs. Instead, the method performs prediction of a hypernym embedding on the basis of a hyponym embedding.  

The contribution of this paper is a novel approach for hypernymy extraction based on projection learning. Namely, we present an improved version of the model proposed by~\newcite{Fu:14}, which makes use of both positive and negative training instances enforcing the asymmetry of the projection. The proposed model is generic and could be straightforwardly used in other relation extraction tasks where both positive and negative training samples are available. Finally, we are the first to successfully apply projection learning for hypernymy extraction in a morphologically rich language. An implementation of our approach and the pre-trained models are available online.\footnote{\url{http://github.com/nlpub/projlearn}}

\section{Related Work}

\textbf{Path-based methods} for hypernymy extraction rely on sentences where both hyponym and hypernym co-occur in characteristic contexts, e.g., ``such \textit{cars} as \textit{Mercedes} and \textit{Audi}''. \newcite{Hearst:92} proposed to use hand-crafted lexical-syntactic patterns to extract hypernyms from such contexts. \newcite{Snow:04} introduced a method for learning patterns automatically based on a set of seed hyponym-hypernym pairs. Further examples of path-based approaches include~\cite{TjongKimSang:09} and \cite{Navigli:10}. The inherent limitation of the path-based methods leading to sparsity issues is that hyponym and hypernym have to co-occur in the same sentence.

Methods based on distributional vectors, such as those generated using the \textit{word2vec} toolkit~\cite{Mikolov:13:w2v}, aim to overcome this sparsity issue as they require no hyponym-hypernym co-occurrence in a sentence. Such methods take representations of individual words as an input to predict relations between them. Two branches of methods relying on distributional representations emerged so far.

\textbf{Methods based on word pair classification} take an ordered pair of word embeddings (a candidate hyponym-hypernym pair) as an input and output a binary label indicating a presence of the hypernymy relation between the words. Typically, a binary classifier is trained on concatenation or subtraction of the input embeddings, cf.~\cite{Roller:14}. Further examples of such methods include~\cite{Lenci:12,Weeds:14,Levy:15,Vylomova:16}. %In particular, \newcite{Vylomova:16} performed a comprehensive evaluation of several approaches for computing semantic relations and found that in word embeddings, vector subtraction generalizes well to a broad range of relations.

HypeNET~\cite{Shwartz:16:hypenet} is a hybrid approach which is also based on a classifier, but in addition to two word embeddings a third vector is used. It represents path-based syntactic information encoded using an LSTM model~\cite{Hochreiter:97}. Their results significantly outperform the ones from previous path-based work of~\newcite{Snow:04}.

An inherent limitation of classification-based approaches is that they require a list of candidate words pairs. While these are given in evaluation datasets such as BLESS~\cite{Baroni:11}, a corpus-wide classification of relations would need to classify all possible word pairs, which is computationally expensive for large vocabularies. Besides, \newcite{Levy:15} discovered a tendency to lexical memorization of such approaches hampering the generalization.

\textbf{Methods based on projection learning} take one hyponym word vector as an input and output a word vector in a topological vicinity of hypernym word vectors. Scaling this to the vocabulary, there is only one such operation per word. \newcite{Mikolov:13:mt} used projection learning for bilingual word translation. \newcite{Vulic:16} presented a systematic study of four classes of methods for learning bilingual embeddings including those based on projection learning. %The approach based on linear projection, similar to the one we use in our method, was found to be the most practical and efficient.

\newcite{Fu:14} were first to apply projection learning for hypernym extraction. Their approach is to learn an affine transformation of a hyponym into a hypernym word vector. The training of their model is performed with stochastic gradient descent. The $k$-means clustering algorithm is used to split the training relations into several groups. One transformation is learned for each group, which can account for the possibility that the projection of the relation depends on a subspace. This state-of-the-art approach serves as the baseline in our experiments.

\newcite{Nayak:15} performed evaluations of distributional hypernym extractors based on classification and projection methods (yet on different datasets, so these approaches are not directly comparable). The best performing projection-based architecture proposed in this experiment is a four-layered feed-forward neural network. No clustering of relations was used. The author used negative samples in the model by adding a regularization term in the loss function. However, drawing negative examples uniformly from the vocabulary turned out to hamper performance. In contrast, our approach shows significant improvements using manually created synonyms and hyponyms as negative samples.

\newcite{yamane-EtAl:2016:COLING} introduced several improvements of the model of~\newcite{Fu:14}. Their model jointly learns projections and clusters by dynamically adding new clusters during training. They also used automatically generated negative instances via a regularization term in the loss function. In contrast to~\newcite{Nayak:15}, negative samples are selected not randomly, but among nearest neighbors of the predicted hypernym. Their approach compares favorably to~\cite{Fu:14}, yet the contribution of the negative samples was not studied. Key differences of our approach from~\cite{yamane-EtAl:2016:COLING} are (1) use of explicit as opposed to automatically generated negative samples, (2) enforcement of asymmetry of the projection matrix via re-projection. While our experiments are based on the model of~\newcite{Fu:14}, our regularizers can be straightforwardly integrated into the model of~\newcite{yamane-EtAl:2016:COLING}. 

\section{Hypernymy Extraction via Regularized Projection Learning}

%Our approach is inspired by additive regularization of topic models~\cite{Vorontsov:15} which lets introducing linguistic constraints to the model by means of regularization.  

\subsection{Baseline Approach}

In our experiments, we use the model of \newcite{Fu:14} as the baseline. In this approach, the projection matrix $\mathbf{\Phi}^*$ is obtained similarly to the linear regression problem, i.e., for the given row word vectors $\vec{x}$ and $\vec{y}$ representing correspondingly hyponym and hypernym, the square matrix $\mathbf{\Phi}^*$ is fit on the training set of positive pairs $\mathcal{P}$:
\begin{equation*}
  \mathbf{\Phi}^* = \arg\min_{\mathbf{\Phi}} \frac{1}{|\mathcal{P}|}
  \sum_{(\vec{x}, \vec{y}) \in \mathcal{P}} \left\|\vec{x}\mathbf{\Phi} - \vec{y}\right\|^2\text{,}
  \label{eq:baseline}
\end{equation*}
where $|\mathcal{P}|$ is the number of training examples and $\|\vec{x}\mathbf{\Phi} - \vec{y}\|$ is the distance between a pair of row vectors $\vec{x}\mathbf{\Phi}$ and $\vec{y}$. In the original method, the $L^2$~distance is used. To improve performance, $k$ projection matrices $\mathbf{\Phi}$ are learned one for each cluster of relations in the training set. One example is represented by a hyponym-hypernym offset. Clustering is performed using the $k$-means algorithm~\cite{MacQueen:67}.

\subsection{Linguistic Constraints via Regularization}

The nearest neighbors generated using distributional word vectors tend to contain a mixture of synonyms, hypernyms, co-hyponyms and other related words~\cite{wandmacher2005semantic,Heylen:08,panchenko:2011:GEMS}. 
In order to explicitly provide examples of undesired relations to the model, we propose two improved versions of the baseline model: \textit{asymmetric regularization} that uses inverted relations as negative examples, and \textit{neighbor regularization} that uses relations of other types as negative examples.  For that, we add a regularization term to the loss function:
\begin{equation*}
  \mathbf{\Phi}^* = \arg\min_{\mathbf{\Phi}} \frac{1}{|\mathcal{P}|}
  \sum_{(\vec{x}, \vec{y}) \in \mathcal{P}} \left\|\vec{x}\mathbf{\Phi} - \vec{y}\right\|^2 + \lambda R\text{,}
  \label{eq:regularized}
\end{equation*}
where $\lambda$ is the constant controlling the importance of the regularization term $R$.

\paragraph{Asymmetric Regularization.} As hypernymy is an asymmetric relation, our first method enforces the asymmetry of the projection matrix. Applying the same transformation to the predicted hypernym vector $\vec{x}\mathbf{\Phi}$ should not provide a vector similar ($\cdot$) to the initial hyponym vector $\vec{x}$. Note that, this regularizer requires only positive examples $\mathcal{P}$:
\begin{equation*}
  R = \frac{1}{|\mathcal{P}|} \sum_{(\vec{x},\_) \in \mathcal{P}} (\vec{x}\mathbf{\Phi}\mathbf{\Phi} \cdot \vec{x})^2.
  \label{eq:hyponym}
\end{equation*}

\vspace{-.75em}\paragraph{Neighbor Regularization.} This approach relies on the negative sampling by explicitly providing the examples of semantically related words $\vec{z}$ of the hyponym $\vec{x}$ that penalizes the matrix to produce the vectors similar to them:
\begin{equation*}
  R = \frac{1}{|\mathcal{N}|} \sum_{(\vec{x}, \vec{z}) \in \mathcal{N}} (\vec{x}\mathbf{\Phi}\mathbf{\Phi} \cdot \vec{z})^2.
  \label{eq:synonym}
\end{equation*}
Note that this regularizer requires negative samples $\mathcal{N}$. In our experiments, we use synonyms of hyponyms as $\mathcal{N}$, but other types of relations can be also used such as antonyms, meronyms or co-hyponyms. Certain words might have no synonyms in the training set. In such cases, we substitute $\vec{z}$ with $\vec{x}$, gracefully reducing to the previous variation. Otherwise, on each training epoch, we sample a random synonym of the given word.

\paragraph{Regularizers without Re-Projection.} In addition to the two regularizers described above, that rely on re-projection of the hyponym vector ($\vec{x}\mathbf{\Phi\Phi}$), we also tested two regularizers without re-projection, denoted as $\vec{x}\mathbf{\Phi}$. The neighbor regularizer in this variation is defined as follows:
\begin{equation*}
  R = \frac{1}{|\mathcal{N}|} \sum_{(\vec{x}, \vec{z}) \in \mathcal{N}} (\vec{x}\mathbf{\Phi} \cdot \vec{z})^2.
  \label{eq:synonymnoreproj}
\end{equation*}
In our case, this regularizer penalizes relatedness of the predicted hypernym $\vec{x}\mathbf{\Phi}$ to the synonym $\vec{z}$. The asymmetric regularizer without re-projection is defined in a similar way. 
\subsection{Training of the Models}

To learn parameters of the considered models we used the Adam method~\cite{Kingma:14} with the default meta-parameters as implemented in the TensorFlow framework~\cite{Abadi:16}.\footnote{\url{https://www.tensorflow.org}} We ran $700$ training epochs passing a batch of $1024$ examples to the optimizer. We initialized elements of each projection matrix using the normal distribution $\mathcal{N}(0, 0.1)$.

\section{Results}

\subsection{Evaluation Metrics}

In order to assess the quality of the model, we adopted the $\hitk{l}$ measure proposed by \newcite{Frome:13} which was originally used for image tagging. For each subsumption pair $(\vec{x}, \vec{y})$ composed of the hyponym $\vec{x}$ and the hypernym $\vec{y}$ in the test set $\mathcal{P}$, we compute $l$ nearest neighbors for the projected hypernym $\vec{x}\mathbf{\Phi}^*$. The pair is considered matched if the gold hypernym $\vec{y}$ appears in the computed list of the $l$ nearest neighbors $\nn{l}(\vec{x}\mathbf{\Phi}^*)$. To obtain the quality score, we average the matches in the test set $\mathcal{P}$:
\begin{equation*}
  \hitk{l} = \frac{1}{|\mathcal{P}|} \sum_{(\vec{x}, \vec{y}) \in \mathcal{P}} \mathbbm{1}\big(
   \vec{y} \in \nn{l}(\vec{x}\mathbf{\Phi}^*) 
  \big)\text{,}
\end{equation*}
where $\mathbbm{1}(\cdot)$ is the indicator function. To consider also the rank of the correct answer, we compute the area under curve measure as the area under the $l-1$ trapezoids:  
$$
\auc = \frac{1}{2} \sum^{l - 1}_{i=1} (\hitk{(i)} + \hitk{(i+1)}).
$$

\begin{figure*}[t]
  \centering
  \includegraphics[width=.475\textwidth]{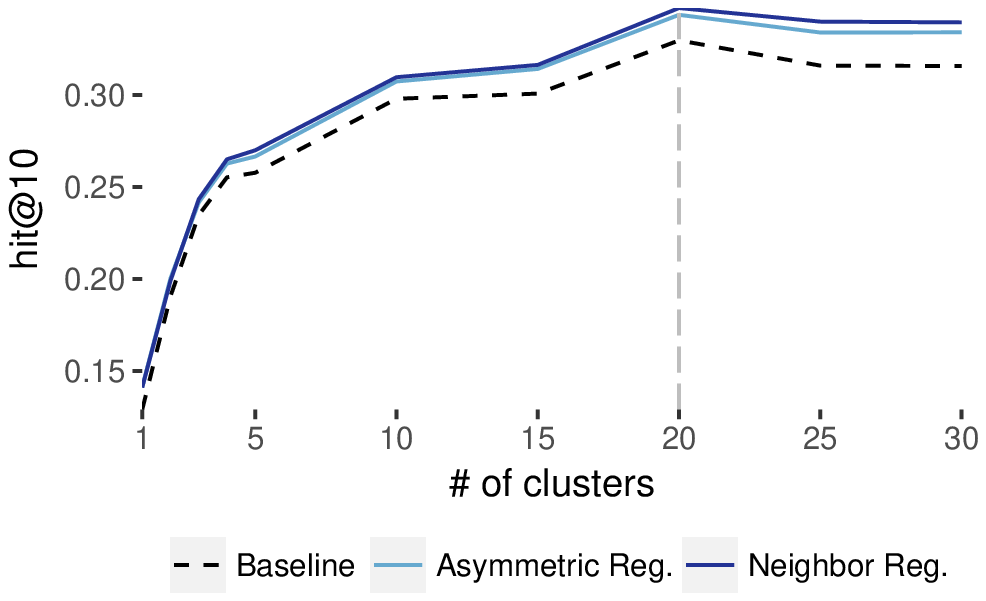} % 420×240
  \quad
  \includegraphics[width=.475\textwidth]{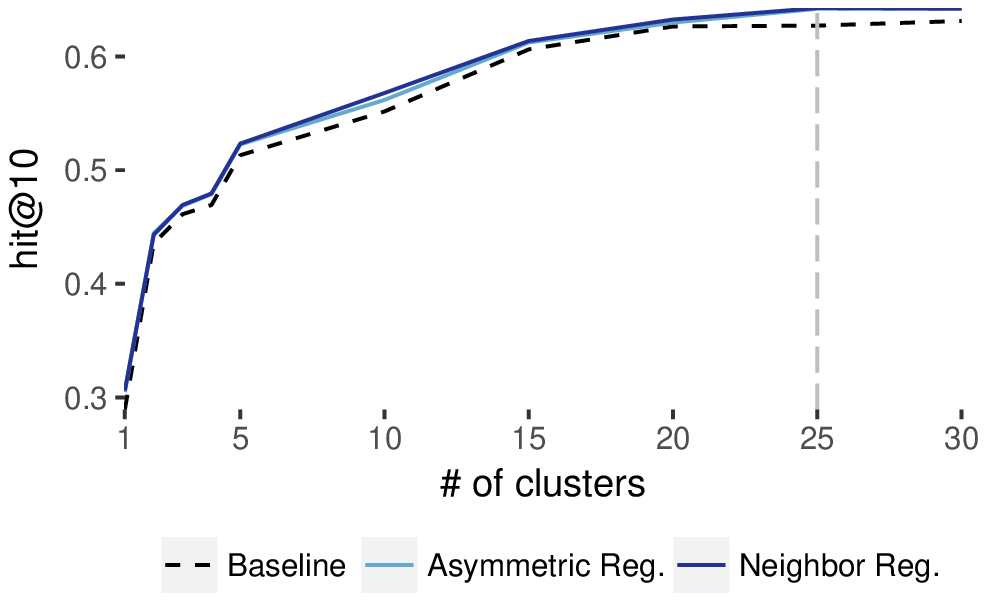} % 380×240
  \vspace{-1em}
  \caption{Performance of our models with re-projection as compared to the baseline approach of~\cite{Fu:14} according to the $\hitk{10}$ measure for Russian (left) and English (right) on the validation set.}
  \label{fig:hit10}
\end{figure*}

\begin{table}[t]
\footnotesize
\centering

\scalebox{0.95}{
\begin{tabular}{ll|cccc}
\textbf{Model}   & & \textbf{hit@1}  &\textbf{hit@5}   & \textbf{hit@10}  & \textbf{AUC}     \\\hline
% ru-sz500-k1-l0.0   baseline                 0.047  0.110  0.130   0.960
% ru-sz500-k1-l0.1   regularized_hyponym_phi  0.059  0.105  0.121   0.926
% ru-sz500-k1-l1.0   regularized_hyponym      0.052  0.120  0.140   1.040
% ru-sz500-k1-l0.1   regularized_synonym_phi  0.059  0.112  0.130   0.980
% ru-sz500-k1-l1.0   regularized_synonym      0.051  0.120  0.139   1.034
% the numbers below are for k=20
Baseline         &                   &
        % $0.209$  &         $0.303$   &         $0.323$  &         $2.665$  \\
        $0.209$  &         $0.303$   &         $0.323$  &         $2.665$  \\
Asym. Reg.  & $\vec{x}\mathbf{\Phi}$     &
        % $0.212$  &         $0.313$   &         $0.333$  &         $2.739$  \\
        $0.213$  &         $0.300$   &         $0.322$  &         $2.659$  \\
Asym. Reg.  & $\vec{x}\mathbf{\Phi\Phi}$ &
        % $0.212$  &         $0.313$   &         $0.333$  &         $2.739$  \\
        $0.212$  &         $0.312$   &         $0.334$  &         $2.743$  \\
Neig. Reg.    & $\vec{x}\mathbf{\Phi}$     &
        % $0.212$  &         $0.313$   &         $0.333$  &         $2.739$  \\
$\mathbf{0.214}$ &         $0.304$   &         $0.325$  &         $2.685$  \\
Neig. Reg.    & $\vec{x}\mathbf{\Phi\Phi}$ &
% $\mathbf{0.212}$ & $\mathbf{0.314}$  & $\mathbf{0.337}$ & $\mathbf{2.765}$ \\
        $0.211$  & $\mathbf{0.315}$  & $\mathbf{0.338}$ & $\mathbf{2.768}$ \\
\end{tabular}
}
\caption{Performance of our approach for Russian for $k=20$ clusters compared to~\cite{Fu:14}.}
\label{tab:performance:ru}
\end{table}

\begin{table*}[t]
\centering
\footnotesize
\begin{tabular}{ll|ccccc|ccccc}
& & & \multicolumn{4}{c|}{EVALution}
    & \multicolumn{5}{c}{EVALution, BLESS, K\&H+N, ROOT09}\\
\textbf{Model} & & $k$ & \textbf{hit@1} & \textbf{hit@5} & \textbf{hit@10} & \textbf{AUC}&
                        $k$ & \textbf{hit@1} & \textbf{hit@5} & \textbf{hit@10} & \textbf{AUC}\\\hline
Baseline         &                   &  $1$ &
        % $0.114$  &          $0.116$  &         $0.121$  &         $1.060$ &
        $0.109$  &          $0.118$  &         $0.120$  &         $1.052$ &
        $1$ & $0.104$  &          $0.247$  &         $0.290$  &         $2.115$\\
Asymmetric Reg.  & $\vec{x}\mathbf{\Phi}$     &  $1$ &
        % $0.134$  &  $\mathbf{0.155}$ &         $0.173$  &         $1.410$ &
        $0.116$  &          $0.125$  &         $0.132$  &         $1.140$ &
        $1$ & $0.132$  &          $0.256$  &         $0.292$  &         $2.204$\\
Asymmetric Reg.  & $\vec{x}\mathbf{\Phi\Phi}$ &  $1$ &
        % $0.134$  &  $\mathbf{0.155}$ &         $0.173$  &         $1.410$ &
        $0.145$  &          $0.166$  &         $0.173$  &         $1.466$ &
        $1$ & $0.112$  &  $\mathbf{0.266}$ &         $0.314$  &         $2.267$\\
Neighbor Reg.    & $\vec{x}\mathbf{\Phi}$     &  $1$ &
% $\mathbf{0.139}$ &  $\mathbf{0.155}$ & $\mathbf{0.177}$ & $\mathbf{1.440}$&
        $0.134$  &          $0.141$  &         $0.150$  &         $1.280$ &
$1$ & $\mathbf{0.134}$ &          $0.255$  &         $0.306$  &         $2.267$ \\
Neighbor Reg.    & $\vec{x}\mathbf{\Phi\Phi}$ &  $1$ &
% $\mathbf{0.139}$ &  $\mathbf{0.155}$ & $\mathbf{0.177}$ & $\mathbf{1.440}$&
$\mathbf{0.148}$ &  $\mathbf{0.168}$ & $\mathbf{0.177}$ & $\mathbf{1.494}$&
$1$ &         $0.111$  &          $0.264$  & $\mathbf{0.316}$ & $\mathbf{2.273}$\\\hline
Baseline         &                   & $30$ &
        % $0.320$  &          $0.339$  &         $0.348$  &         $3.040$ &
        $0.327$  &          $0.339$  &         $0.350$  &         $3.080$ &
$25$ &         $0.546$  &          $0.614$  &         $0.634$  &         $5.481$\\
Asymmetric Reg.  & $\vec{x}\mathbf{\Phi}$     & $30$ &
        % $0.339$  &          $0.355$  &         $0.366$  &         $3.210$ &
        $0.336$  &          $0.354$  &         $0.366$  &         $3.201$ &
$25$ &         $0.547$  &          $0.616$  &         $0.632$  &         $5.492$\\
Asymmetric Reg.  & $\vec{x}\mathbf{\Phi\Phi}$ & $30$ &
        % $0.339$  &          $0.355$  &         $0.366$  &         $3.210$ &
        $0.341$  &          $0.364$  &         $0.368$  &         $3.255$ &
$25$ & $\mathbf{0.553}$ &          $0.621$  & $\mathbf{0.642}$ &         $5.543$\\
Neighbor Reg.    & $\vec{x}\mathbf{\Phi}$     & $30$ &
% $\mathbf{0.341}$ &  $\mathbf{0.361}$ & $\mathbf{0.377}$ & $\mathbf{3.270}$&
        $0.339$  &          $0.357$  &         $0.364$  &         $3.210$ &
$25$ &         $0.547$  &          $0.617$  &         $0.634$  &         $5.494$ \\
Neighbor Reg.    & $\vec{x}\mathbf{\Phi\Phi}$ & $30$ &
% $\mathbf{0.341}$ &  $\mathbf{0.361}$ & $\mathbf{0.377}$ & $\mathbf{3.270}$&
$\mathbf{0.345}$ &  $\mathbf{0.366}$ & $\mathbf{0.370}$ & $\mathbf{3.276}$&
$25$ &  $\mathbf{0.553}$ &  $\mathbf{0.623}$ &         $0.641$  & $\mathbf{5.547}$\\
\end{tabular}
\caption{Performance of our approach for English without clustering $(k=1)$ and with the optimal number of cluster on the EVALution datasets ($k=30$) and on the combined datasets ($k=25$). 
% we do not need "opt": we have two k values (see table)
}
\label{tab:performance:en}
\vspace{-1.25em}
\end{table*}

\subsection{Experiment 1: The Russian Language}

\paragraph{Dataset.} In this experiment, we use word embeddings published as a part of the Russian Distributional Thesaurus~\cite{Panchenko:16} trained on $12.9$ billion token collection of Russian books. The embeddings were trained using the skip-gram model~\cite{Mikolov:13:w2v} with $500$ dimensions and a context window of $10$ words.

The dataset used in our experiments has been composed of two sources. We extracted synonyms and hypernyms from the Wiktionary\footnote{\url{http://www.wiktionary.org}} using the Wikokit toolkit~\cite{Krizhanovsky:13}. To enrich the lexical coverage of the dataset, we extracted additional hypernyms from the same corpus using Hearst patterns for Russian using the PatternSim toolkit~\cite{Panchenko:12}.\footnote{\url{https://github.com/cental/patternsim}} To filter noisy extractions, we used only relations extracted more than $100$ times.

As suggested by~\newcite{Levy:15}, we split the train and test sets such that each contains a distinct vocabulary to avoid the lexical overfitting. This results in $25\,067$ training, $8\,192$ validation, and $8\,310$ test examples. The validation and test sets contain hypernyms from Wiktionary, while the training set is composed of hypernyms and synonyms coming from both sources.

\paragraph{Discussion of Results.}

\figurename~\ref{fig:hit10} (left) shows performance of the three projection learning setups on the validation set: the baseline approach, the asymmetric regularization approach, and the neighbor regularization approach. Both regularization strategies lead to consistent improvements over the non-regularized baseline of~\cite{Fu:14} across various cluster sizes. The method reaches optimal performance for $k=20$ clusters. \tablename~\ref{tab:performance:ru} provides a detailed comparison of the performance metrics for this setting. Our approach based on the regularization using synonyms as negative samples outperform the baseline (all differences between the baseline and our models are significant with respect to the $t$-test). According to all metrics, but $\hitk{1}$ for which results are comparable to $\vec{x}\mathbf{\Phi}$, the re-projection ($\vec{x}\mathbf{\Phi\Phi}$) improves results. 

\subsection{Experiment 2: The English Language}

We performed the evaluation on two datasets.

\paragraph{EVALution Dataset.} In this evaluation, word embeddings were trained on a $6.3$ billion token text collection composed of Wikipedia, ukWaC~\cite{Ferraresi:08}, Gigaword~\cite{Graff:03}, and news corpora from the Leipzig Collection \cite{Goldhahn:12}. We used the skip-gram model with the context window size of $8$ tokens and $300$-dimensional vectors.

We use the EVALution dataset~\cite{Santus:15} for training and testing the model, composed of $1\,449$ hypernyms and $520$ synonyms, where hypernyms are split into $944$ training, $65$ validation and $440$ test pairs. Similarly to the first experiment, we extracted extra training hypernyms using the Hearst patterns, but in contrast to Russian, they did not improve the results significantly, so we left them out for English. A reason for such difference could be the more complex morphological system of Russian, where each word has more morphological variants compared to English. Therefore, extra training samples are needed for Russian (embeddings of Russian were trained on a non-lemmatized corpus).

\paragraph{Combined Dataset.} To show the robustness of our approach across configurations, this dataset has more training instances, different embeddings, and both synonyms and co-hyponyms as negative samples. We used  hypernyms, synonyms and co-hyponyms from the four commonly used datasets: EVALution, BLESS~\cite{Baroni:11}, ROOT09~\cite{Santus:16} and K\&H+N~\cite{Necsulescu:15}.%\footnote{\url{https://github.com/vered1986/LexNET}}
The obtained $14\,528$ relations were split into $9\,959$ training, $1\,631$ validation and $1\,625$ test hypernyms; $1\,313$ synonyms and co-hyponyms were used as negative samples. We used the standard $300$-dimensional embeddings trained on the $100$ billion tokens Google News corpus~\cite{Mikolov:13:w2v}.%\footnote{\url{https://code.google.com/archive/p/word2vec}}

\paragraph{Discussion of Results.} \figurename~\ref{fig:hit10} (right) shows that similarly to Russian, both regularization strategies lead to consistent improvements over the non-regularized baseline. \tablename~\ref{tab:performance:en} presents detailed results for both English datasets. Similarly to the first experiment, our approach consistently improves results robustly across various configurations. As we change the number of clusters, types of embeddings, the size of the training data and type of relations used for negative sampling, results using our method stay superior to those of the baseline. The regularizers without re-projection ($\vec{x}\mathbf{\Phi}$) obtain lower results in most configurations as compared to re-projected versions ($\vec{x}\mathbf{\Phi\Phi}$). Overall, the neighbor regularization yields slightly better results in comparison to the asymmetric regularization. We attribute this to the fact that some synonyms $\vec{z}$ are close to the original hyponym $\vec{x}$, while others can be distant. Thus, neighbor regularization is able to safeguard the model during training from more errors. This is also a likely reason why the performance of both regularizers is similar: the asymmetric regularization makes sure that a re-projected vector does not belong to a semantic neighborhood of the hyponym. Yet, this is exactly what neighbor regularization achieves. Note, however that neighbor regularization requires explicit negative examples, while asymmetric regularization does not.

\section{Conclusion}

In this study, we presented a new model for extraction of hypernymy relations based on the projection of distributional word vectors. The model incorporates information about explicit negative training instances represented by relations of other types, such as synonyms and co-hyponyms, and enforces asymmetry of the projection operation. Our experiments in the context of the hypernymy prediction task for English and Russian languages show significant improvements of the proposed approach over the state-of-the-art model without negative sampling. 

\section*{Acknowledgments}

We acknowledge the support of the Deutsche For\-schungs\-gemeinschaft (DFG) foundation under the ``JOIN-T'' project, the Deutscher Akademischer Austauschdienst (DAAD), the Russian Foundation for Basic Research (RFBR) under the project no.~16-37-00354 mol\_a, and the Russian Foundation for Humanities under the project no.~16-04-12019 ``RussNet and YARN thesauri integration''. We also thank Microsoft for providing computational resources under the Microsoft Azure for Research award. Finally, we are grateful to Benjamin Milde, Andrey Kutuzov, Andrew Krizhanovsky, and Martin Riedl for discussions and suggestions related to this study.

\bibliography{eacl2017}

\begin{thebibliography}{}

\bibitem[\protect\citename{Abadi~et al.}2016]{Abadi:16}
Mart{\'{\i}}n Abadi~et al.
\newblock 2016.
\newblock {TensorFlow: Large-Scale Machine Learning on Heterogeneous
  Distributed Systems}.
\newblock {\em CoRR}, abs/1603.04467.

\bibitem[\protect\citename{Baroni and Lenci}2011]{Baroni:11}
Marco Baroni and Alessandro Lenci.
\newblock 2011.
\newblock {How We BLESSed Distributional Semantic Evaluation}.
\newblock In {\em Proceedings of the GEMS 2011 Workshop on GEometrical Models
  of Natural Language Semantics}, GEMS '11, pages 1--10, Edinburgh, Scotland.
  Association for Computational Linguistics.

\bibitem[\protect\citename{Ferraresi \bgroup et al.\egroup }2008]{Ferraresi:08}
Adriano Ferraresi, Eros Zanchetta, Marco Baroni, and Silvia Bernardini.
\newblock 2008.
\newblock {Introducing and evaluating ukWaC, a very large Web-derived corpus of
  English}.
\newblock In {\em Proceedings of the 4th Web as Corpus Workshop (WAC-4): Can we
  beat Google?}, pages 47--54, Marakech, Morocco.

\bibitem[\protect\citename{Frome \bgroup et al.\egroup }2013]{Frome:13}
Andrea Frome, Greg~S. Corrado, Jon Shlens, Samy Bengio, Jeff Dean,
  Marc'~Aurelio Ranzato, and Tomas Mikolov.
\newblock 2013.
\newblock {DeViSE: A Deep Visual-Semantic Embedding Model}.
\newblock In {\em Advances in Neural Information Processing Systems 26}, pages
  2121--2129. Curran Associates, Inc., Harrahs and Harveys, NV, USA.

\bibitem[\protect\citename{Fu \bgroup et al.\egroup }2014]{Fu:14}
Ruiji Fu, Jiang Guo, Bing Qin, Wanxiang Che, Haifeng Wang, and Ting Liu.
\newblock 2014.
\newblock {Learning Semantic Hierarchies via Word Embeddings}.
\newblock In {\em Proceedings of the 52nd Annual Meeting of the Association for
  Computational Linguistics (Volume~1: Long Papers)}, pages 1199--1209,
  Baltimore, MD, USA. Association for Computational Linguistics.

\bibitem[\protect\citename{Goldhahn \bgroup et al.\egroup }2012]{Goldhahn:12}
Dirk Goldhahn, Thomas Eckart, and Uwe Quasthoff.
\newblock 2012.
\newblock {Building Large Monolingual Dictionaries at the Leipzig Corpora
  Collection: From 100 to 200 Languages}.
\newblock In {\em Proceedings of the Eight International Conference on Language
  Resources and Evaluation (LREC'12)}, pages 759--765, Istanbul, Turkey.
  European Language Resources Association (ELRA).

\bibitem[\protect\citename{Gong \bgroup et al.\egroup }2005]{Gong:05}
Zhiguo Gong, Chan~Wa Cheang, and U.~Leong~Hou.
\newblock 2005.
\newblock {Web Query Expansion by WordNet}.
\newblock In {\em Proceedings of the 16th International Conference on Database
  and Expert Systems Applications - DEXA '05}, pages 166--175. Springer Berlin
  Heidelberg, Copenhagen, Denmark.

\bibitem[\protect\citename{Graff}2003]{Graff:03}
David Graff.
\newblock 2003.
\newblock {English Gigaword}.
\newblock Technical Report LDC2003T05, Linguistic Data Consortium,
  Philadelphia, PA, USA.

\bibitem[\protect\citename{Hearst}1992]{Hearst:92}
Marti~A. Hearst.
\newblock 1992.
\newblock {Automatic Acquisition of Hyponyms from Large Text Corpora}.
\newblock In {\em Proceedings of the 14th Conference on Computational
  Linguistics - Volume~2}, COLING'92, pages 539--545, Nantes, France.
  Association for Computational Linguistics.

\bibitem[\protect\citename{Heylen \bgroup et al.\egroup }2008]{Heylen:08}
Kris Heylen, Yves Peirsman, Dirk Geeraerts, and Dirk Speelman.
\newblock 2008.
\newblock {Modelling Word Similarity: an Evaluation of Automatic Synonymy
  Extraction Algorithms}.
\newblock In {\em Proceedings of the Sixth International Conference on Language
  Resources and Evaluation (LREC'08)}, pages 3243--3249, Marrakech, Morocco.
  European Language Resources Association (ELRA).

\bibitem[\protect\citename{Hochreiter and Schmidhuber}1997]{Hochreiter:97}
Sepp Hochreiter and J\"{u}rgen Schmidhuber.
\newblock 1997.
\newblock {Long Short-Term Memory}.
\newblock {\em Neural Computation}, 9(8):1735--1780.

\bibitem[\protect\citename{Kingma and Ba}2014]{Kingma:14}
Diederik~P. Kingma and Jimmy Ba.
\newblock 2014.
\newblock {Adam: A Method for Stochastic Optimization}.
\newblock {\em CoRR}, abs/1412.6980.

\bibitem[\protect\citename{Krizhanovsky and Smirnov}2013]{Krizhanovsky:13}
Andrew~A. Krizhanovsky and Alexander~V. Smirnov.
\newblock 2013.
\newblock {An approach to automated construction of a general-purpose lexical
  ontology based on Wiktionary}.
\newblock {\em Journal of Computer and Systems Sciences International},
  52(2):215--225.

\bibitem[\protect\citename{Lenci and Benotto}2012]{Lenci:12}
Alessandro Lenci and Giulia Benotto.
\newblock 2012.
\newblock {Identifying Hypernyms in Distributional Semantic Spaces}.
\newblock In {\em Proceedings of the First Joint Conference on Lexical and
  Computational Semantics - Volume~1: Proceedings of the Main Conference and
  the Shared Task, and Volume~2: Proceedings of the Sixth International
  Workshop on Semantic Evaluation}, SemEval '12, pages 75--79, Montr\'{e}al,
  Canada. Association for Computational Linguistics.

\bibitem[\protect\citename{Levy \bgroup et al.\egroup }2015]{Levy:15}
Omer Levy, Steffen Remus, Chris Biemann, and Ido Dagan.
\newblock 2015.
\newblock {Do Supervised Distributional Methods Really Learn Lexical Inference
  Relations?}
\newblock In {\em Proceedings of the 2015 Conference of the North American
  Chapter of the Association for Computational Linguistics: Human Language
  Technologies}, pages 970--976, Denver, Colorado, USA. Association for
  Computational Linguistics.

\bibitem[\protect\citename{MacQueen}1967]{MacQueen:67}
James MacQueen.
\newblock 1967.
\newblock {Some methods for classification and analysis of multivariate
  observations}.
\newblock In {\em Proceedings of the Fifth Berkeley Symposium on Mathematical
  Statistics and Probability, Volume~1: Statistics}, pages 281--297, Berkeley,
  California, USA. University of California Press.

\bibitem[\protect\citename{Mikolov \bgroup et al.\egroup }2013a]{Mikolov:13:mt}
Tomas Mikolov, Quoc~V. Le, and Ilya Sutskever.
\newblock 2013a.
\newblock {Exploiting Similarities among Languages for Machine Translation}.
\newblock {\em CoRR}, abs/1309.4168.

\bibitem[\protect\citename{Mikolov \bgroup et al.\egroup
  }2013b]{Mikolov:13:w2v}
Tomas Mikolov, Ilya Sutskever, Kai Chen, Greg~S. Corrado, and Jeffrey Dean.
\newblock 2013b.
\newblock {Distributed Representations of Words and Phrases and their
  Compositionality}.
\newblock In {\em Advances in Neural Information Processing Systems 26}, pages
  3111--3119. Curran Associates, Inc., Harrahs and Harveys, NV, USA.

\bibitem[\protect\citename{Miller}1995]{Miller:95}
George~A. Miller.
\newblock 1995.
\newblock {WordNet: A Lexical Database for English}.
\newblock {\em Communications of the ACM}, 38(11):39--41.

\bibitem[\protect\citename{Navigli and Velardi}2010]{Navigli:10}
Roberto Navigli and Paola Velardi.
\newblock 2010.
\newblock {Learning Word-Class Lattices for Definition and Hypernym
  Extraction}.
\newblock In {\em Proceedings of the 48th Annual Meeting of the Association for
  Computational Linguistics}, pages 1318--1327, Uppsala, Sweden. Association
  for Computational Linguistics.

\bibitem[\protect\citename{Nayak}2015]{Nayak:15}
Neha Nayak.
\newblock 2015.
\newblock {Learning Hypernymy over Word Embeddings}.
\newblock Technical report, Stanford University.

\bibitem[\protect\citename{Necsulescu \bgroup et al.\egroup
  }2015]{Necsulescu:15}
Silvia Necsulescu, Sara Mendes, David Jurgens, N\'{u}ria Bel, and Roberto
  Navigli.
\newblock 2015.
\newblock {Reading Between the Lines: Overcoming Data Sparsity for Accurate
  Classification of Lexical Relationships}.
\newblock In {\em Proceedings of the Fourth Joint Conference on Lexical and
  Computational Semantics}, pages 182--192, Denver, CO, USA. Association for
  Computational Linguistics.

\bibitem[\protect\citename{Panchenko \bgroup et al.\egroup }2012]{Panchenko:12}
Alexander Panchenko, Olga Morozova, and Hubert Naets.
\newblock 2012.
\newblock {A Semantic Similarity Measure Based on Lexico-Syntactic Patterns}.
\newblock In {\em Proceedings of KONVENS 2012}, pages 174--178, Vienna,
  Austria. \"{O}GAI.

\bibitem[\protect\citename{Panchenko \bgroup et al.\egroup
  }2016a]{panchenko-EtAl:2016:SemEval}
Alexander Panchenko, Stefano Faralli, Eugen Ruppert, Steffen Remus, Hubert
  Naets, Cedrick Fairon, Simone~Paolo Ponzetto, and Chris Biemann.
\newblock 2016a.
\newblock {TAXI at SemEval-2016 Task 13: a Taxonomy Induction Method based on
  Lexico-Syntactic Patterns, Substrings and Focused Crawling}.
\newblock In {\em Proceedings of the 10th International Workshop on Semantic
  Evaluation (SemEval-2016)}, pages 1320--1327, San Diego, CA, USA. Association
  for Computational Linguistics.

\bibitem[\protect\citename{Panchenko \bgroup et al.\egroup
  }2016b]{Panchenko:16}
Alexander Panchenko, Dmitry Ustalov, Nikolay Arefyev, Denis Paperno, Natalia
  Konstantinova, Natalia Loukachevitch, and Chris Biemann.
\newblock 2016b.
\newblock {Human and Machine Judgements for Russian Semantic Relatedness}.
\newblock In {\em Proceedings of the 5th Conference on Analysis of Images,
  Social Networks and Texts (AIST'2016)}, volume 661 of {\em Communications in
  Computer and Information Science}, pages 303--317, Yekaterinburg, Russia.
  Springer-Verlag Berlin Heidelberg.

\bibitem[\protect\citename{Panchenko}2011]{panchenko:2011:GEMS}
Alexander Panchenko.
\newblock 2011.
\newblock {Comparison of the Baseline Knowledge-, Corpus-, and Web-based
  Similarity Measures for Semantic Relations Extraction}.
\newblock In {\em Proceedings of the GEMS 2011 Workshop on GEometrical Models
  of Natural Language Semantics}, pages 11--21, Edinburgh, UK. Association for
  Computational Linguistics.

\bibitem[\protect\citename{Roller \bgroup et al.\egroup }2014]{Roller:14}
Stephen Roller, Katrin Erk, and Gemma Boleda.
\newblock 2014.
\newblock {Inclusive yet Selective: Supervised Distributional Hypernymy
  Detection}.
\newblock In {\em Proceedings of COLING 2014, the 25th International Conference
  on Computational Linguistics: Technical Papers}, pages 1025--1036, Dublin,
  Ireland, August. Dublin City University and Association for Computational
  Linguistics.

\bibitem[\protect\citename{Santus \bgroup et al.\egroup }2015]{Santus:15}
Enrico Santus, Frances Yung, Alessandro Lenci, and Chu-Ren Huang.
\newblock 2015.
\newblock {EVALution 1.0: an Evolving Semantic Dataset for Training and
  Evaluation of Distributional Semantic Models}.
\newblock In {\em Proceedings of the 4th Workshop on Linked Data in
  Linguistics: Resources and Applications}, pages 64--69, Beijing, China.
  Association for Computational Linguistics.

\bibitem[\protect\citename{Santus \bgroup et al.\egroup }2016]{Santus:16}
Enrico Santus, Alessandro Lenci, Tin-Shing Chiu, Qin Lu, and Chu-Ren Huang.
\newblock 2016.
\newblock {Nine Features in a Random Forest to Learn Taxonomical Semantic
  Relations}.
\newblock In {\em Proceedings of the Tenth International Conference on Language
  Resources and Evaluation (LREC 2016)}, pages 4557--4564, Portoro\v{z},
  Slovenia. European Language Resources Association (ELRA).

\bibitem[\protect\citename{Shwartz \bgroup et al.\egroup
  }2016]{Shwartz:16:hypenet}
Vered Shwartz, Yoav Goldberg, and Ido Dagan.
\newblock 2016.
\newblock {Improving Hypernymy Detection with an Integrated Path-based and
  Distributional Method}.
\newblock In {\em Proceedings of the 54th Annual Meeting of the Association for
  Computational Linguistics (Volume~1: Long Papers)}, pages 2389--2398, Berlin,
  Germany. Association for Computational Linguistics.

\bibitem[\protect\citename{Snow \bgroup et al.\egroup }2004]{Snow:04}
Rion Snow, Daniel Jurafsky, and Andrew~Y. Ng.
\newblock 2004.
\newblock {Learning Syntactic Patterns for Automatic Hypernym Discovery}.
\newblock In {\em Proceedings of the 17th International Conference on Neural
  Information Processing Systems}, NIPS'04, pages 1297--1304, Vancouver,
  British Columbia, Canada. MIT Press.

\bibitem[\protect\citename{Snow \bgroup et al.\egroup }2006]{Snow:06}
Rion Snow, Daniel Jurafsky, and Andrew~Y. Ng.
\newblock 2006.
\newblock {Semantic Taxonomy Induction from Heterogenous Evidence}.
\newblock In {\em Proceedings of the 21st International Conference on
  Computational Linguistics and 44th Annual Meeting of the Association for
  Computational Linguistics}, pages 801--808, Sydney, Australia. Association
  for Computational Linguistics.

\bibitem[\protect\citename{Tjong Kim~Sang and Hofmann}2009]{TjongKimSang:09}
Erik Tjong Kim~Sang and Katja Hofmann.
\newblock 2009.
\newblock {Lexical Patterns or Dependency Patterns: Which Is Better for
  Hypernym Extraction?}
\newblock In {\em Proceedings of the Thirteenth Conference on Computational
  Natural Language Learning (CoNLL-2009)}, pages 174--182, Boulder, Colorado,
  USA. Association for Computational Linguistics.

\bibitem[\protect\citename{Vuli\'{c} and Korhonen}2016]{Vulic:16}
Ivan Vuli\'{c} and Anna Korhonen.
\newblock 2016.
\newblock {On the Role of Seed Lexicons in Learning Bilingual Word Embeddings}.
\newblock In {\em Proceedings of the 54th Annual Meeting of the Association for
  Computational Linguistics (Volume~1: Long Papers)}, pages 247--257, Berlin,
  Germany. Association for Computational Linguistics.

\bibitem[\protect\citename{Vylomova \bgroup et al.\egroup }2016]{Vylomova:16}
Ekaterina Vylomova, Laura Rimell, Trevor Cohn, and Timothy Baldwin.
\newblock 2016.
\newblock {Take and Took, Gaggle and Goose, Book and Read: Evaluating the
  Utility of Vector Differences for Lexical Relation Learning}.
\newblock In {\em Proceedings of the 54th Annual Meeting of the Association for
  Computational Linguistics (Volume~1: Long Papers)}, pages 1671--1682, Berlin,
  Germany. Association for Computational Linguistics.

\bibitem[\protect\citename{Wandmacher}2005]{wandmacher2005semantic}
Tonio Wandmacher.
\newblock 2005.
\newblock {How semantic is Latent Semantic Analysis?}
\newblock In {\em Proceedings of R\'{E}CITAL 2005}, pages 525--534, Dourdan,
  France.

\bibitem[\protect\citename{Weeds \bgroup et al.\egroup }2014]{Weeds:14}
Julie Weeds, Daoud Clarke, Jeremy Reffin, David Weir, and Bill Keller.
\newblock 2014.
\newblock {Learning to Distinguish Hypernyms and Co-Hyponyms}.
\newblock In {\em Proceedings of COLING 2014, the 25th International Conference
  on Computational Linguistics: Technical Papers}, pages 2249--2259, Dublin,
  Ireland. Dublin City University and Association for Computational
  Linguistics.

\bibitem[\protect\citename{Yamane \bgroup et al.\egroup
  }2016]{yamane-EtAl:2016:COLING}
Josuke Yamane, Tomoya Takatani, Hitoshi Yamada, Makoto Miwa, and Yutaka Sasaki.
\newblock 2016.
\newblock {Distributional Hypernym Generation by Jointly Learning Clusters and
  Projections}.
\newblock In {\em Proceedings of COLING 2016, the 26th International Conference
  on Computational Linguistics: Technical Papers}, pages 1871--1879, Osaka,
  Japan, December. The COLING 2016 Organizing Committee.

\bibitem[\protect\citename{Zhou \bgroup et al.\egroup }2013]{Zhou:13}
Guangyou Zhou, Yang Liu, Fang Liu, Daojian Zeng, and Jun Zhao.
\newblock 2013.
\newblock {Improving Question Retrieval in Community Question Answering Using
  World Knowledge}.
\newblock In {\em Proceedings of the Twenty-Third International Joint
  Conference on Artificial Intelligence}, IJCAI '13, pages 2239--2245, Beijing,
  China. AAAI Press.

\end{thebibliography}
\bibliographystyle{eacl2017}

\end{document}